\documentclass{article}

\usepackage{arxiv}

\usepackage[utf8]{inputenc} 
\usepackage[T1]{fontenc}    
\usepackage{hyperref}       
\usepackage{url}            
\usepackage{booktabs}       
\usepackage{amsfonts}       
\usepackage{nicefrac}       
\usepackage{microtype}      
\usepackage{lipsum}
\usepackage{graphicx}
\graphicspath{ {./images/} }

\usepackage{algpseudocode}
\usepackage{algorithm}

\title{On the impact of interpretability methods in Active Image Augmentation method}

\author{
 Flávio Arthur Oliveira Santos \\
  Centro de Informática\\
  Universidade Federal de Pernambuco\\
  Recife, Brazil \\
  \texttt{faos@cin.ufpe.br} \\
   \And
 Cleber Zanchettin \\
  Centro de Informática\\
  Universidade Federal de Pernambuco\\
  Recife, Brazil \\
  \texttt{cz@cin.ufpe.br} \\
  \And
 Leonardo Matos \\
  Departamento de Ciência da Computação\\
  Universidade Federal de Sergipe\\
  São Cristóvão, Brazil \\
  \texttt{leonardo@dcomp.ufs.br} \\
  \And
 Paulo Novais \\
  Informatics Department \\
  University of Minho\\
  Braga, Portugal \\
  \texttt{pjon@di.uminho.pt} \\
}

\begin{document}

\maketitle

\begin{abstract}
Robustness is a significant constraint in machine learning models. The performance of the algorithms must not deteriorate when training and testing with slightly different data. Deep neural network models achieve awe-inspiring results in a wide range of applications of computer vision. Still, in the presence of noise or region occlusion, some models exhibit inaccurate performance even with data handled in training. Besides, some experiments suggest deep learning models sometimes use incorrect parts of the input information to perform inference. Activate Image Augmentation (ADA) is an augmentation method that uses interpretability methods to augment the training data and improve its robustness to face the described problems. Although ADA presented interesting results, its original version only used the Vanilla Backpropagation interpretability to train the U-Net model. In this work, we propose an extensive experimental analysis of the interpretability method's impact on ADA. We use five interpretability methods: Vanilla Backpropagation, Guided Backpropagation, GradCam, Guided GradCam, and InputXGradient. The results show that all methods achieve similar performance at the ending of training, but when combining ADA with GradCam, the U-Net model presented an impressive fast convergence.    
\end{abstract}


\section{Introduction}
Deep Learning (DL) \cite{deeplearning} models have achieved state-of-the-art results in a great number of different tasks. For example, image segmentation \cite{Kirillov_2020_CVPR}, visual question $\&$ answering \cite{Jiang_2020_CVPR} and super-resolution \cite{superresolution}. In computer vision tasks, the Convolutional Neural Network (CNN) \cite{lecun1995convolutional} has stood out and presented excellent results. Although DL achieves excellent results, the way how the models are trained makes them a black box, so turning it difficult to explain "why" the model made some inference. In the DL literature, there are some approaches to visualize the most important input information to network inference, for example, Vanilla Backpropagation \cite{saliency} and Guided Backpropagation methods\cite{guidedbackprop}.

\begin{figure}[h]
    \centering
    \includegraphics[width=\linewidth]{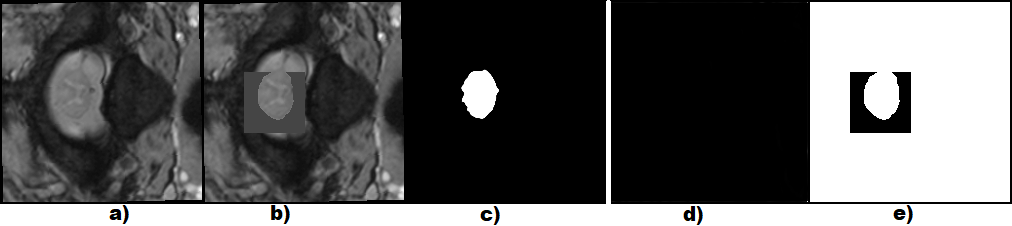}
    \caption{a) Original input image; b) Edited input image to remove contextual pixels; c) Original segmentation (considering original image); d) New segmentation (considering edited image); e) Mask used to edit the input image. The mask is used to change the values of contextual pixels of the target information. After removing those pixels the model is not abble to perform the segmentation. Image adapted from \cite{ada}.}
    \label{fig:unet-err}
\end{figure}

In addition to the interpretability problem, there is no guarantee that the DL model inference is based on information that the user considers essential to the related task (target problem). In \cite{ada}, we presented an example where a U-Net model \cite{unet} is used to perform gray matter segmentation \cite{spinalchallenge}. The model achieved a Dice score \cite{dice} of 0.91 on the validation set, which is an excellent result considering related works \cite{deepseg}. However, excluding some pixels of the input image (not related to gray matter area), the U-Net model can no longer segment the input image correctly, even having all the gray matter information (target problem) in the image. The Figure~\ref{fig:unet-err} illustrates this situation. This behavior is a robustness problem because even keeping the total gray matter area the input the model does not segment the image correctly. As suggested in \cite{ada}, the U-Net model is using contextual pixels and not only the gray matter pixels (target problem) to perform segmentation.

Active Image Augmentation (ADA) \cite{ada} is a method created to mitigate the problem above. ADA augment the training data set adding new images with some pixels set to 0. Thus, forcing the model to self-adapt and consider the pixels considered important for the target problem (pixel with values different from 0). For choosing what pixels set to 0 in each image, ADA uses the Vanilla Gradient \cite{saliency} representation to obtain the degree of importance of each input image pixel and remove some contextual pixels of this region. The area of pixels with information considered crucial is not changed (e.g., gray matter region). Although ADA has achieved excellent robustness results in Spinal Cord Grey Matter Segmentation (SCGM)  tasks \cite{spinalchallenge}, the original paper \cite{ada} only investigates the Vanilla Backpropagation \cite{saliency} method to identify the occlusion region. In this paper, we perform an experimental analysis of the ADA method to evaluate the impact of the interpretability methods on its approach. We compare the performance of Vanilla Backpropagation to other well-known interpretability methods such as GradCam \cite{gradcam}, Guided Backpropagation \cite{guidedbackprop}, Guided Gradcam \cite{gradcam}, and InputXGradient \cite{inputxgradient}. 

The remaining of the paper is structured as follows: Section 2 presents the related works. In Section 3, we discuss the Active Image Augmentation method, while Section 4 and 5 show the experimental analysis and obtained results, respectively. The conclusions are presented in Section 6.

\section{Interpretability Methods}
The Vanilla Backpropagation \cite{saliency} method aims to build a saliency map of the input image relating each pixel's importance to the model inference. Given an input image $x$ to the deep learning model, we can obtain a score $c$ of the class activation inferred by the model. The Vanilla Gradient computes the gradient of the score $c$ with relation to the input $x$ to obtain the saliency map. DeconvNet \cite{zeiler2014visualizing} is an interpretability method based on the Deconvolutional Networks approach \cite{zeiler2010deconvolutional}. It computes the gradient of the output unit c with relation to the input image to build the saliency maps, however, it applies a ReLU in the gradients of all ReLU layers output. 

Guided Backpropagation \cite{guidedbackprop} is similar to Vanilla Backprop \cite{saliency}. However, in the process of building the saliency maps, it only considers the intermediate gradient values whose values are positives. Thus,  the generated saliency maps are cleaner than Vanilla Backprop. Input X Gradient \cite{inputxgradient} is a straightforward approach; it consists of basically compute the gradient of a given class $c$ with relation to the input and then multiply it by $x$ input vector. 

Gradient-Weighted Class Activation Mapping (Grad-Cam) \cite{gradcam} uses the gradient information to produce the activation maps highlighting the essential parts to the model prediction. Different from the methods presented earlier, DeConvNet and Guided Backpropagation, Grad-Cam can highlight the most discriminative information to the predicted class. In contrast, the other methods obtain high-resolution maps highlighting many details in the input image. In a general way, Grad-Cam computes the weighted gradient of the output $c$ with relation to the feature map $A_{k}$. In the following, it applies ReLU \cite{relupaper} in the linear combination results to obtain just the positives contributions. Although Grad-Cam finds class-discriminative regions in the input vector, the model fails to identify important details in the pixel space. 

\section{Review of Active Image Data Augmentation}
Active Image Data Augmentation (ADA) \cite{ada} is an approach to improve model robustness and automatically guide the model to focus on the most important region of the input features. It consists of generating new training data in each training cycle. To produce new training data, the ADA uses the interpretability method to identify where the deep learning model is focusing on the input vector. Next, it selects the most important region of $N \times N$ dimension automatically in the input vector and set all the pixel values of this region to zero. After applying ADA systematically, the model will learn to focus on feature regions related to the target problem. 

The equations \ref{rna}-\ref{xnew} where obtained from \cite{ada} and represent the ADA method step-by-step. \\[-10mm]

\begin{equation}
    y = f(x;\theta)
    \label{rna}
\end{equation}

Equation~\ref{rna} represents a model with a input $x$ and the parameters set $\theta$. \\[-8mm]

\begin{equation}
    maps(i, x) = \left\Vert \frac{\partial y[i]}{\partial x} \right\Vert
    \label{gradientMaps}
\end{equation}

The equation~\ref{gradientMaps} represents a step of building the saliency maps of the output model with relation to the input vector. In this example the interpretability method is the Vanilla Backprop \cite{saliency}, but different interpretability methods can be used. \\[-10mm]

\begin{equation}
    mask = build\_mask(maps(i, x), ground\_truth, n, z)
    \label{mask}
\end{equation}


\begin{equation}
    x\_new = x * mask
    \label{xnew}
\end{equation}

The function $build\_mask$ (Equation~\ref{mask}) build a binary mask of $N \times N$ dimension. Only the contiguous region of dimension $Z \times Z$ has value 0, all other pixels are set to 1. The $Z \times Z$ region is the most important information to the model inference considering on the interpretability method. The $*$ operation in equation \ref{xnew} is a point-wise multiplication between two vectors. The resulting $x\_new$ vector is the new input data close to $x$ but with the most import $Z \times Z$ region removed. The ADA \cite{ada} presents an algorithm detailing how to implement the $build\_mask$ function in a computationally efficient way. 

It is important to note that the Active Image Data Augmentation method differs from removing some input data information because it chose what to remove based on the interpretability methods. Thus, it is a driven way of data augmentation.

\subsection{Active Image Data Augmentation Training}
The Algorithm \ref{alg:2} presents the ADA training method. It consists of two major steps: (i) training the model during \textit{$standard\_epochs$} using the original data; and (ii) execute several cycles where each cycle generates new data from ADA and train the model during $ada\_epochs$.

The ADA training method execute $(standard\_epochs + cycles*ada\_epochs)$ training epochs. This number of epochs is important to evaluate the computational cost of ADA.

\begin{algorithm}[H]
\caption{ADA training method. Adapted from \cite{ada}.}
\label{alg:2}

\begin{algorithmic}[1]
\Function{$Ada\_Training$}{$model, data, cycles, ada\_epochs$}
    \State $conventional\_data \gets data + classic\_augmentations(data)$
    \For{$i \gets 0; i < standard\_epochs$}
        \State $model.train(conventional\_data)$
    \EndFor
    \For{$i \gets 0; i < cycles$}
        \State $new\_data \gets conventional\_data + Ada(data, model)$
        \For{$j \gets 0; j < ada\_epochs$} 
            \State $model.train(new\_data)$
        \EndFor
    \EndFor
    \State \Return $model$
\EndFunction
\end{algorithmic}
\end{algorithm}

\section{Experiments}
We investigate five different interpretability methods to evaluate its impact on model robustness. The evaluated methods are Vanilla Gradient, Guided Backpropagation, GradCam, Guided Gradcam, and InputXGradient. Following the same setup as ADA, we used the spinal cord grey matter (SCGM) segmentation task as the reference data set. We used the same U-Net \cite{unet} approach presented on \cite{ada}. The following sections describe the SCGM data set, evaluation metrics, U-Net architecture, and details about analysis.

\subsection{Spinal Cord Grey Matter Segmentation}
The spinal cord grey matter segmentation (SCGM) challenge released a data set composed of magnetic resonance imaging (MRI) data of different subjects. All challenge data comprises 80 datasets, grouped in 40 for training and 40 for the test. According to SCGM \cite{spinalchallenge}, each 20 data set was acquired at one of the following sites: University College London, Polytechnique Montreal, University of Zurich, and Vanderbilt University. Specific details about the data can be found in \cite{spinalchallenge}. To produce the validation data, we used $20\%$ of the original training data. Our experiment has a robustness evaluation scenario, which is very important to our approach. We used the validation data to build this robustness data because we do not have access to the SCGM test set. The evaluation process with the test set is performed in an online system \footnote{http://niftyweb.cs.ucl.ac.uk/program.php?p=CHALLENGE}.

\subsection{Metrics}
We used five different metrics to evaluate the models. These metrics measure different types of information between the model output and ground truth. The metrics are divided into overlap, distance, and statistical approaches. Table \ref{tab:metrics} presents the names, abbreviations, range, and category of all metrics. 

\begin{table}[H]
\centering
\caption{Resume of the evaluation metrics. Adapted from \cite{ada}.}
\label{tab:metrics}
\setlength{\tabcolsep}{8pt} 
\renewcommand{\arraystretch}{1.3} 
\begin{tabular}{l|c|c|c}
\hline
\textbf{Metric Name} &  \textbf{Abbr.} & \textbf{Range} & \textbf{Category} \\
\hline
Dice Similarity Coefficient              &    DSC    &  $0–100$    &    Overlap \\
Hausdorff Surface Distance               &    HSD    &  $>0$     &    Distance \\
Sensitivity (TP)         &  TPR &  $0–100$  & Statistical \\
Specificity (TN)         &  TNR &  $0–100$  & Statistical \\
Precision   &    PPV &  $0–100$    & Statistical \\
\hline
\end{tabular}
\end{table}

The lower values in the distance metrics mean better results, but higher values represent better results in overleaped and statistical metrics. The following list clarifies the meaning of some results. Given a ground truth (GT) mask and the mask obtained by neural network model output (MO), a voxel can be considered:
 
\begin{itemize}
     \item True Positive (TP): classified as GM voxel in GT and MO;
     \item True Negative (TN): classified as non-GM voxel in GT and MO;
     \item False Positive (FP): classified as non-GM voxel in GT but was classified as GM voxel in MO;
     \item False Negative (FN): classified as GM voxel in GT but was classified as non-GM voxel in MO.
 \end{itemize}
 
\subsection{Models and Training Details}
Since the model choice is an important step in this work, the same U-Net architecture suggested in \cite{ada} was used in experiments. The U-Net comprises three downsample layers, 3 UpSamples layers, and a logistic sigmoid activation function in the output layer. It is important to highlight that after every convolutional layer has a dropout layer and a batch normalization step. During the first 100 training epochs, the following traditional data augmentation methods where employed: Rotation, Shift, Scale, Chanel shift, and Elastic deformation. Table \ref{tab:augmentation} presents all the used parameters defined based on previous works \cite{perone2018spinal}.
  
The U-Net was trained during 100 epochs with only the original data. Then, 5 U-Net models were created, initialized with the U-Net state at epoch 100. Each model was trained using the ADA with a specific interpretability method: (i) Vanilla Backprop; (ii) InputXGradient; (iii) GradCam; (iv) Guided Backprop; e (v) Guided GradCam. We decide to use those interpretability methods because they have the same characteristics. All of them are gradient-based and do not need a baseline input, such as Integrated Gradients \cite{integratedgradients}. Such as in \cite{ada}, the U-Net uses a dropout rate of $0.5$ and a batch normalization momentum of $0.4$. The optimization method is Adam \cite{adam}, with an initial learning rate of $0.001$ and a batch size $16$. Every U-Net model was trained during $31$ ADA cycles, where each cycle training during $30$ epochs with data from ADA using occlusion region of $20 \times 20$. To produce the robustness results, we chose the epoch of each model with the best validation results. 

\begin{table}[H]
\centering
\caption{Traditional data augmentation parameters.}
\label{tab:augmentation}
\setlength{\tabcolsep}{8pt} 
\renewcommand{\arraystretch}{1.3} 
\begin{tabular}{l|c}
\hline
\textbf{Method}       & \textbf{Parameter} \\
\hline
Rotation (degrees)    & $[-4.6, 4.6]$      \\
Shift $(\%)$          & $[-0.03, 0.03]$    \\
Scaling               & $[0.98, 1.02]$     \\
Channel Shift         & $[-0.17, 0.17]$    \\
Elastic Deformation   & $\alpha = 30.0, \sigma = 4.0$    \\
\hline
\end{tabular}
\end{table}
 
\subsection{Robustness Data Set}
To build the robustness dataset, we used the validation data from SCGM. We generate new images with a region of 20x20, occluding parts of each image. Thus, we evaluate the model performance in a scenario with images containing incomplete information. The Algorithm \ref{alg:1} shows how we build the robustness data set. 

\begin{algorithm}[!h]
\caption{How to build the robustness data function.}
\label{alg:1}

\begin{algorithmic}[1]
\Function{$Build\_Robustness\_Data$}{$data, n, z$}
    \State $robustness\_data \gets []$
    \For {$x \in data$}    
        \For{$i \gets 0; i < n; i += z$}
            \For{$j \gets 0; j < n; j += z$} 
                \State $x\_occluded \gets erase\_region(clone(x), i, j, z)$
                \State $robustness\_data.append(x\_occluded)$
            \EndFor
        \EndFor
    \EndFor
    \State \Return $robustness\_data$
\EndFunction
\end{algorithmic}
\end{algorithm}

The $erase\_region$ function in Algorithm \ref{alg:1} is a function that receives a input image and set to 0 all its pixels in the square of size $z^2$ at position $(i, j)$ and finishing at position $(i + z, j + z)$.

\section{Results and Discussions}

\begin{figure}[h]
    \centering
    \includegraphics[scale=0.4]{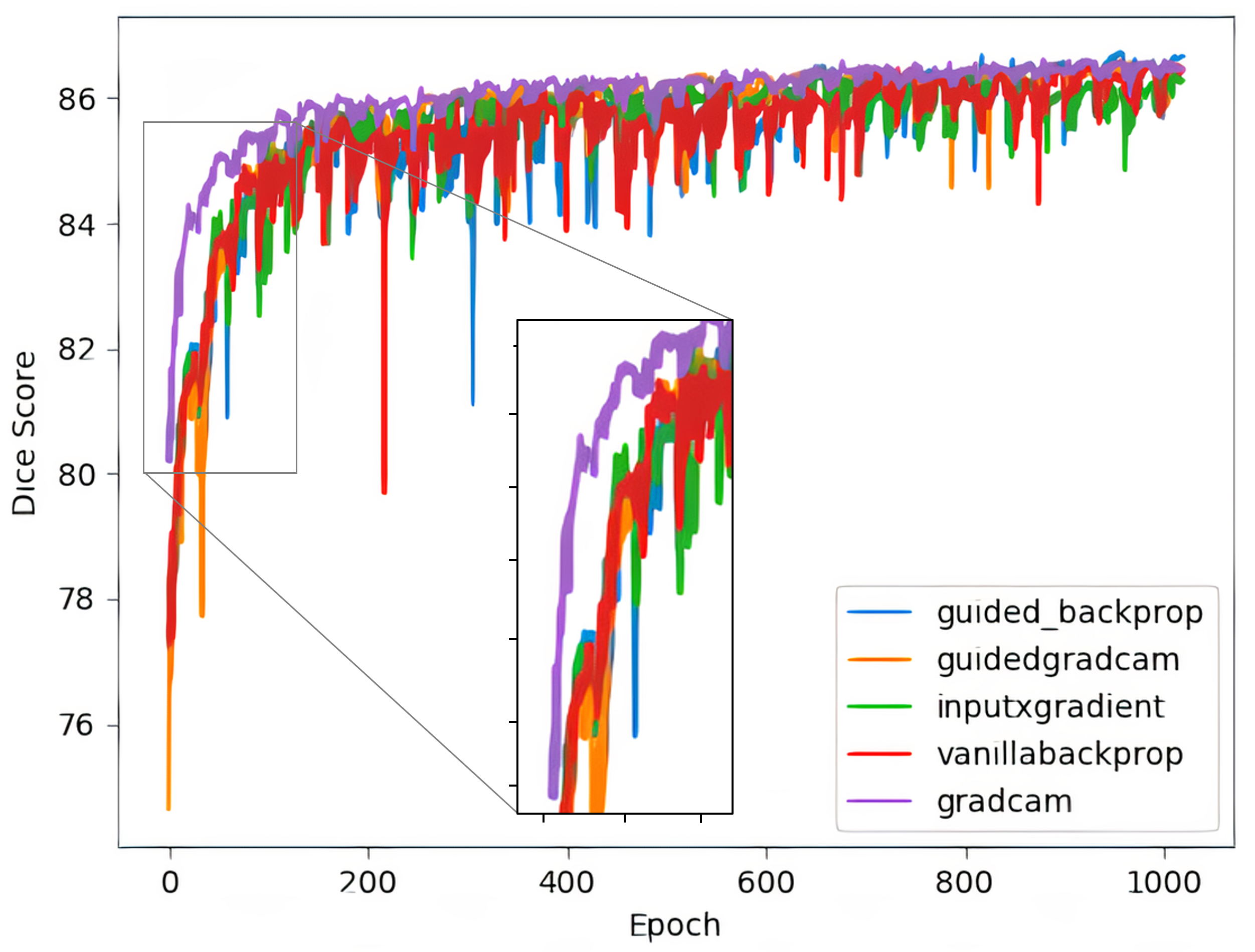}
    \caption{Analysis of the model convergence. The zoom area shows the best performance of GradCam in the first training epochs.}
    \label{fig:analysis}
\end{figure}

Figure \ref{fig:analysis} presents the convergence curve of the five U-Net models, each one is using a different interpretability method. This convergence curve was produced by evaluating the U-Net models in each epoch with original validation data. The epoch analysis shows that the U-Net using GradCam has an impressive validation result in the first epoch, exceeding a dice score of 80, while all other methods present a dice score smaller than 79. This result can be evidence that before applying the ADA method, the U-Net model was using the wrong information to perform the inference process. However, after the first epoch of ADA using GradCam, the model could change the point of interest to the possibly right information, thus reaching an exciting validation result after just the first epoch. In addition to the very impressive starting point of the GradCam, the curve shows that it overcomes all other models at epoch 123, then all methods converge to almost the same point and present similar performance. 

\begin{table}[h]
\centering
\caption{Results obtained using the validation data from SCGM challenge.}
\label{tab:results-validation}
\setlength{\tabcolsep}{8pt} 
\renewcommand{\arraystretch}{1.3} 
\begin{tabular}{l|c|c|c|c|c}
\hline
\textbf{Models}  & \textbf{DSC} & \textbf{HSD} & \textbf{TPR} & \textbf{TNR} & \textbf{PPV} \\
\hline
Vanilla Backprop    & 86.56          & 1.72          & 87.72          & 99.95 & 86.21 \\
InputXGradient      & 86.45          & \textbf{1.70} & 87.33          & 99.95 & 86.38 \\
Guided Gradient     & \textbf{86.72} & 1.71          & 87.57          & 99.95 & \textbf{86.70} \\
GradCam             & 86.61          & 1.72          & \textbf{89.53} & 99.94 & 84.61 \\
Guided GradCam      & 86.56          & \textbf{1.70} & 87.65          & 99.95 & 86.34 \\
\hline
\end{tabular}
\end{table}

The main goal of ADA is to improve the robustness of the model. Thus, a robustness data set was created from the validation data to evaluate the five trained models. The details about how the robustness data set was built are present in section 4.4. All results presented so far using validation data are important only to show that the models trained using the ADA method keep the original inference performance. Table \ref{tab:results-robustness} present the results obtained using the robustness data. The results show that the U-Net with GradCam presents better results than all other models. Besides, the precision-recall behavior of U-Net with GradCam is different from all other models, being precision slightly lower than others and recall higher. That results shows that the ADA method present impressive results on robustness using all five interpretability methods. Furthermore, the result achieved by the GradCam method presents strong evidence that the GradCam method is capable of guiding the U-Net to focus on the real important information. 

\begin{table}[h]
\centering
\caption{Results obtained using the robustness data.}
\label{tab:results-robustness}
\setlength{\tabcolsep}{8pt} 
\renewcommand{\arraystretch}{1.3} 
\begin{tabular}{l|c|c|c|c|c}
\hline
\textbf{Models}  & \textbf{DSC} & \textbf{HSD} &  \textbf{TPR} & \textbf{TNR} & \textbf{PPV} \\
\hline
Vanilla Backprop    & 85.86          & 1.72          & 86.73          & 99.94 & 85.99  \\
InputXGradient      & 85.76          & 1.71          & 86.36          & 99.95 & 86.15  \\
Guided Gradient     & 86.03          & 1.71          & 86.57          & 99.95 & \textbf{86.50}  \\
GradCam             & \textbf{86.15} & 1.73          & \textbf{88.84} & 99.94 & 84.45  \\
Guided GradCam      & 85.90          & \textbf{1.70} & 86.68          & 99.95 & 86.12  \\
\hline
\end{tabular}
\end{table}


After the analysis of figure \ref{fig:analysis}, the table \ref{tab:results-validation} highlight the results of each model on the validation data. Every model presents a similar best dice score (DSC), but the U-Net model using Guided Gradient presents the better DSC, followed by U-Net using GradCam. It is worth highlighting the difference in precision and recall displayed by the GradCam model compared to all other models; in future works this behavior should be further investigated. By combining Figure 2 and the results presented in Table 3, it is known that, in the latest epochs, each model has reached almost the same quality. Therefore, the differences that have an impact were found in the first cycles of ADA methods.

\subsection{Analysis of the First Augmentation Steps}
From the obtained results, we show that the differences in the impact among the interpretability methods are found in the initial stages of the ADA execution. Thus, in this section, we compare the interpretations obtained by the different methods in the first step of ADA. 

\begin{figure}[h]
    \centering
    \includegraphics[width=\textwidth]{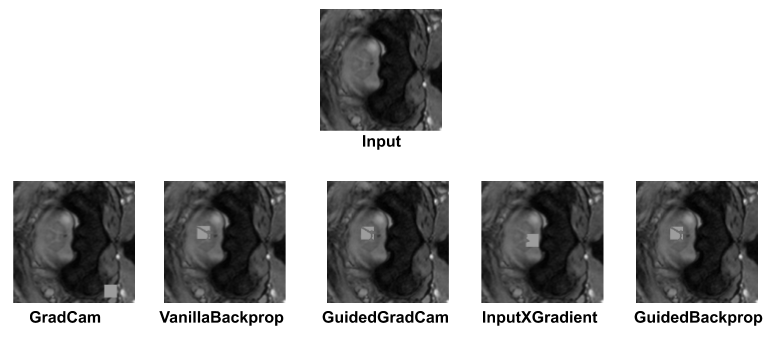}
    \caption{Augmentation sample obtained in the first ADA cycle. Given an input image, we compute the IoU score between the regions that each method assigns as important.}
    \label{fig:blursample}
\end{figure}

For each input image in the training data, the interpretability method employed on ADA suggests which region of the input we must occlude to build the augmented data training. Thus, given an image, we compute the IoU score between the regions that each method assigns as important to obtain the similarity between them. Figure \ref{fig:blursample} presents a sample of the original input and the augmented images obtained from the interpretability methods. The IoU is computed based on the mask used to occlude each region of the input image. 

\begin{figure}[h!]
    \centering
    \includegraphics[scale=0.30]{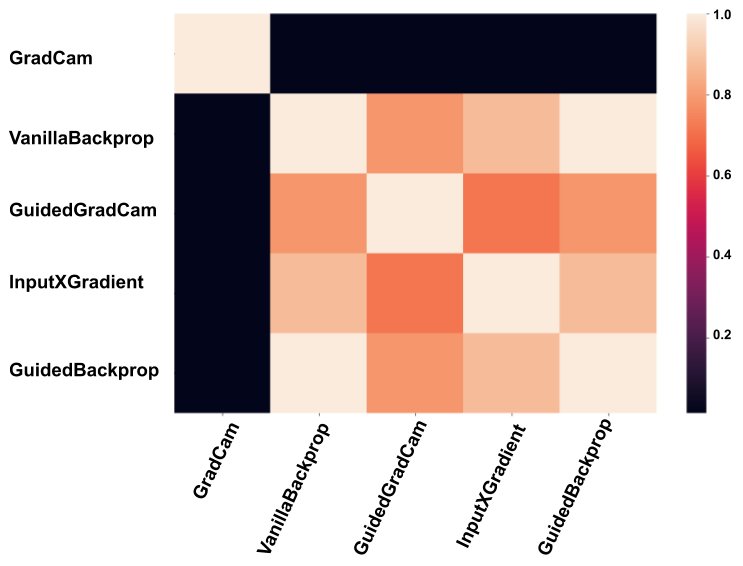}
    \caption{IoU matrix of all methods. The GradCam has a very low IoU compared to all other methods. This implies that the regions occluded from GradCam are very different from all other approaches, as present in Figure \ref{fig:blursample}.}
    \label{fig:iou-analysis}
\end{figure}

The IoU matrix in Figure \ref{fig:iou-analysis} present the average IoU computed from all of the images in the training dataset. Since this IoU matrix is obtained in the first step of ADA and the GradCam presented the better results in the first epoch, this experiment demonstrates that the GradCam produces very different training images from all other models.

\section{Conclusion}
Active Image Data augmentation is a method to improve the model robustness helping the model focus on the important input information. We evaluate different interpretability methods on ADA to identify its influence on the approach performance. 

The results show that all the interpretability methods achieved competitive results, without losing the performance obtained when training with the original data. In addition, the robustness scenario shows that all the models achieved similar results in the last epochs, but in the first epochs, the approach of ADA using GradCam presented impressive results. This suggests that GradCam guided ADA to help the U-Net model deal with the right input information early on the first epochs. 

Based on the convergence curve of all models, they have similar results. As future work, the same performance of the different methods in the last epochs should be investigated. Presumably, it all produces very similar interpretations in the last training epochs.

\section*{Acknowledgments}

This work has been supported by FCT - Fundação para a Ciência e Tecnologia within the Project Scope: UIDB/00319/2020. The authors also thanks CAPES and CNPq (Brazilian Researcher Agencies) for the finantial support.

\bibliographystyle{unsrt}  
\bibliography{references}  


\end{document}